\documentclass[letterpaper]{article}
\usepackage{aaai2026}
\usepackage{times}
\usepackage{helvet}
\usepackage{courier}
\usepackage[hyphens]{url}
\usepackage{graphicx}
\usepackage{natbib}
\usepackage{caption}
\usepackage{amsmath}
\usepackage{amssymb}
\usepackage{booktabs}
\usepackage{dblfloatfix}
\usepackage[dvipsnames,table,HTML]{xcolor}
\usepackage{listings}
\usepackage{algorithm}
\usepackage{algorithmic}
\title{Robust Multi-Agent Path Finding under Observation Attacks: \\
A Principled Adversarial-Plus-Smoothing Training Recipe}
\author{Riad Ahmed}
\affiliations{University of New Hampshire\\riad.ahmed@unh.edu\\[4pt]
\textbf{Project page:} \url{https://riadahmedunh.github.io/Robust_MAPF_page/}}

\begin{document}
\maketitle

% =====================================================================
\begin{abstract}
% =====================================================================
Decentralized multi-agent path finding (MAPF) routes a team of agents
on a shared grid, each acting from its own local view. The standard
solution trains one shared neural policy with Proximal Policy
Optimization (PPO), a popular on-policy reinforcement learning
algorithm. Such a policy works well on clean observations, but a small
input perturbation on one agent often changes its action, which then
blocks a neighbour, and the team jams. In this paper we present two
training recipes that keep the same network and the same deployment
loop, yet make the policy hold up under perturbed observations. The
first recipe, Adv-PPO, trains the shared policy against worst-case
perturbations of its own input and selects the checkpoint by
performance under adversarial perturbation. The second recipe, Adv-PPO+MACER, fine-tunes
that checkpoint with a small on-policy smoothness term whose gradient
follows the certified radius of randomized smoothing. On POGEMA with
$8{\times}8$ maps and four agents, the unprotected PPO policy reaches
$95.8\%$ clean success but only $2.5\%$ under the strongest attack.
Adv-PPO recovers worst-case success to $59.2\%$ at one percentage
point of clean cost. Adv-PPO+MACER recovers it to $77.5\%\pm 6.0\%$
across three independent seeds at less than one percentage point of
clean cost. We support these numbers with per-attack curves, a
certified action-stability sanity check (which measures the
smoothed-policy wrapper, not the deployed argmax policy), and
side-by-side rollout storyboards that show the failure mode and the
fix inside one environment instance.
\end{abstract}

% =====================================================================
\section{Introduction}
% =====================================================================

A delivery robot, a warehouse forklift, and a stocking shuttle share
the same aisle. None of them sees the whole floor. Each one sees only
a small window around itself and decides where to step next. Multi-agent
path finding (MAPF) is the formal version of this problem: route $N$
agents from start cells to goal cells on a shared grid without
collisions \citep{stern2019mapfdef}. When agents act with only a local
view, the standard solution is to train one neural network with
Proximal Policy Optimization (PPO) \citep{ppo}, share its parameters
across all agents, and let each agent pick its own action from its own
window. This shared-policy template is the basis for PRIMAL
\citep{primal} and for recent benchmarks such as POGEMA \citep{pogema}
and Follower \citep{follower}, and it reaches high clean success rates
on standard maps.

The trouble starts when the input is no longer clean. A real sensor
sometimes reports an obstacle that is not there; upstream perception
modules misfire on the wrong cell for all sorts of reasons. More
directly, an attacker can craft a small input bump that makes the
policy pick a wrong action. Because all
agents share the same network, the same wrong action is selected
everywhere. One robot turns into the path of another, the other has to
wait, the team collapses around the disturbance. We measure this on
POGEMA $8{\times}8$ with four agents: a single sign-gradient step
\citep{goodfellow_fgsm} on the local observation tensor at
$\ell_\infty$ budget $\epsilon=0.20$ takes the same policy from
$95.8\%$ success on clean inputs down to $2.5\%$ success in the worst
attacked cell (Table~\ref{tab:headline}). The fragility is not a
corner case. It is what happens when a deployed policy meets imperfect
sensors.

We want to keep the parts of the standard pipeline that already work:
the same shared-parameter network, the same PPO outer loop, and the same
deployment loop with no extra runtime component. We do not add a denoiser
in front of the sensor, and we do not change the backbone architecture.
The only place we make changes is the training objective itself.

This paper proposes two training recipes that work in that space, both
built from published robustness ingredients but combined and tuned for
shared-policy MAPF. The first recipe is \textbf{Adv-PPO}. During
training we attack the policy's own observation with a frozen
pre-trained baseline, add a regularizer from state-adversarial PPO
\citep{sa-mdp} that asks the policy to act consistently across nearby
inputs, add a small local-smoothness term in the spirit of TRADES
\citep{trades}, and pick the released checkpoint by performance on a
held-out attacked validation set instead of by clean reward. The
second recipe is \textbf{Adv-PPO+MACER}. We start from the strongest
Adv-PPO checkpoint and add a small on-policy term whose gradient is
the gradient of the certified-radius bound of randomized smoothing
\citep{cohen2019,macer}. The MACER hinge is paired with a slightly
larger entropy bonus so that the policy keeps exploring instead of
collapsing onto a single deterministic action.

Both recipes recover most of the lost robustness with very small
clean cost. Adv-PPO lifts worst-case attacked success from $2.5\%$ to
$59.2\%$ while clean success moves from $95.8\%$ to $96.7\%$.
Adv-PPO+MACER lifts the worst-case to $77.5\%\pm 6.0\%$ across three
independent fine-tune seeds while clean success stays at
$95.0\%\pm 0.8\%$. A paired bootstrap over the $21$ attack settings
gives a $+2.4$ percentage-point gain of Adv-PPO+MACER over Adv-PPO on
mean attacked success with $95\%$ confidence interval $[+0.4,+4.9]$,
so the certified-radius hinge gives a real cell-level improvement on
top of the SA-PPO recipe.

We report two recipes because they sit at different cost-quality
points. Adv-PPO is simpler, has fewer hyperparameters, and is the
right starting point for a practitioner who wants robustness with the
least added machinery. Adv-PPO+MACER is the stronger of the two and
is the configuration we recommend when the additional fine-tune cost
is acceptable. Reporting both also lets the reader see what the
certified-radius hinge adds on top of state-adversarial training.

\paragraph{Contributions.}
\begin{enumerate}
\item \textbf{Adv-PPO}: a shared-policy adversarial training recipe for
decentralized MAPF that lifts worst-case attacked success on POGEMA
$8{\times}8$ from $2.5\%$ to $59.2\%$ with no clean-performance cost
(Sec.~\ref{sec:advppo}).
\item \textbf{Adv-PPO+MACER}: an on-policy integration of the MACER
certified-radius hinge into the PPO loop, which further lifts worst-case
success to $77.5\%\pm 6.0\%$ across three independent seeds while clean
success stays at $95.0\%\pm 0.8\%$ (Sec.~\ref{sec:macer}).
\item A finding on when certified-radius training helps and when it
hurts: the same hinge applied as a post-hoc distillation step collapses
deployed clean success, while the on-policy version does not. We
explain why and draw a practical lesson for integrating certified-radius
signals into reinforcement learning (RL) training.
\item A reproducible evaluation covering five types of attacks, $21$
attack configurations, three independent seeds, and rollout
visualizations that show what the numbers look like inside a real episode.
\end{enumerate}

% =====================================================================
\section{Related Work}
% =====================================================================

\paragraph{Decentralized MAPF.} Centralized MAPF planners such as
Conflict-Based Search \citep{cbs} solve the same problem optimally
under full observability and no per-agent compute budget; the
decentralized RL setting we study trades that optimality for local
observability, real-time per-agent decisions, and graceful degradation
when an agent's input is corrupted. PRIMAL \citep{primal} introduced
shared-policy reinforcement learning for MAPF. POGEMA \citep{pogema}
and Follower \citep{follower} provide modern benchmarks; we use POGEMA
as is and do not modify the environment.

\paragraph{State-adversarial RL.} \citet{sa-mdp} formulate the
state-adversarial Markov Decision Process (SA-MDP) and propose SA-PPO
(state-adversarial PPO) and SA-DQN (state-adversarial Deep Q-Network).
Their experiments
are single-agent control. We extend the regularizer to a shared-policy
multi-agent setting and combine it with a smoothness penalty and an
on-policy MACER hinge. \citet{rarl} co-train a destabilizing adversary
alongside the protagonist; \citet{action_robust} show that a fixed
perturbation source is already useful as a regularizer. Our
training-time attacks come from a frozen pre-trained baseline, in line
with the latter argument.

\paragraph{Adversarial examples and certified robustness.}
Fast Gradient Sign Method (FGSM)
\citep{goodfellow_fgsm} and Projected Gradient Descent (PGD)
\citep{madry_pgd} are the two attacks
we test. TRADES \citep{trades} introduces a clean-vs-robust trade-off
through an inner Kullback-Leibler (KL) term; our smoothness penalty is a uniform-noise
relative of TRADES that drops the inner PGD step. Randomized smoothing
\citep{cohen2019} gives a probabilistic certified radius for any
classifier. MAximum CERtified Radius (MACER) \citep{macer} turns this radius into a differentiable
training loss, which we re-use inside PPO with the modifications we
describe in Sec.~\ref{sec:macer}.

\paragraph{Robust cooperative MARL.} \citet{lin2020robust} attack
cooperative agents in StarCraft II. \citet{guo2022adversarial} survey
attacks and defenses in MARL. \citet{adv_policies} show that adversarial
\emph{policies} can defeat self-play agents, which is a different
threat model from ours. We study the simpler case of per-agent
observation perturbations on a shared decentralized policy, so the
baseline and the gain are both easy to read.

% =====================================================================
\section{Problem Formulation}
\label{sec:problem}
% =====================================================================

We work in the standard partially-observed cooperative MAPF setting on
POGEMA, then specify the threat model and the metrics we report.

\paragraph{Environment and policy.} A POGEMA episode is parameterized
by a grid of side $L$, an obstacle density $\rho$, and a team of $N$
agents. Each agent $i$ has a start cell $s^{(i)}_0$ and a goal cell
$g^{(i)}$. At step $t$, agent $i$ receives a local observation
\[
o^{(i)}_t \in \{0,1\}^{C\times(2r+1)\times(2r+1)},
\]
which is a window of radius $r$ around its current cell. The three
channels encode obstacles, the positions of other agents, and a single
hint cell that points toward $g^{(i)}$. The action set is
$\mathcal{A}=\{\textsc{wait},\textsc{up},\textsc{down},\textsc{left},
\textsc{right}\}$. All agents share one policy
$\pi_\theta(\cdot\mid o)$ with parameters $\theta$. The episode ends
when all agents reach their goals or the horizon $T$ is reached.

\paragraph{Reward.} The training reward gives $+1$ when an agent
reaches its goal, a small per-step time penalty, and a small collision
penalty. The number that matters at evaluation time is success rate,
defined below.

\paragraph{Threat model.} The physical grid, obstacles, and goal positions are
fixed for the duration of an episode. Each episode is drawn from a random seed,
so different seeds give different map layouts and different starting positions for
the agents; the variation we study is across seeds, not within one episode.
What varies within an episode is the observation each agent receives: the attacker
injects a small perturbation $\delta^{(i)}_t \in \mathcal{D}(o^{(i)}_t,\epsilon)$ at
every step, where
\begin{equation}
\begin{aligned}
\mathcal{D}(o,\epsilon) = \{\delta:\,&\|\delta\|_\infty\le\epsilon,\\[-1pt]
&o+\delta\in[0,1]^{C\times(2r+1)\times(2r+1)}\},
\end{aligned}
\label{eq:budget}
\end{equation}
and the policy sees $\tilde{o}^{(i)}_t = o^{(i)}_t + \delta^{(i)}_t$.
The attacker's objective is to make the policy stop preferring the
clean-input action:
\begin{equation}
\delta^{(i)}_t \in \arg\!\!\max_{\delta\in\mathcal{D}(o^{(i)}_t,\epsilon)}\!\!
\ell\!\left(\pi_\phi(\cdot\mid o^{(i)}_t+\delta),\;
a^{(i),\text{clean}}_t\right),
\label{eq:attacker}
\end{equation}
where $\phi$ are the parameters of the source network used to compute
the attack (the defender during evaluation, the frozen baseline during
training; see Sec.~\ref{sec:method}), and $\ell$ is the cross-entropy
between the policy distribution and the one-hot clean action
$a^{(i),\text{clean}}_t = \arg\max_a \pi_\phi(a\mid o^{(i)}_t)$. The
$\ell_\infty$ ball with radius $\epsilon$ is the standard model for
small bounded sensor errors and is the same model used by
\citet{sa-mdp}. Because $o\in\{0,1\}$, the per-pixel ball
$[\max(0,o-\epsilon),\min(1,o+\epsilon)]$ is asymmetric: an
informative occupied pixel can only be attacked downward by
$\epsilon$, an empty pixel can only be attacked upward by $\epsilon$.
This makes the attacker strictly weaker than an unconstrained
$\ell_\infty$ attack on a real-valued tensor, and we account for it in
our reported numbers (we always clip to $[0,1]$).

\paragraph{Concrete attacks.} FGSM is one sign-gradient step at full
$\epsilon$ and a clip to $\mathcal{D}$. PGD is $K$ smaller sign steps
of size $\alpha_{\text{step}}=2\epsilon/K$ from one random initialization
in $\mathcal{D}$, projected back to $\mathcal{D}$ at each step (no
additional restarts). Sensor-noise attacks add Gaussian noise,
salt-and-pepper noise, or per-channel dropout to $o^{(i)}_t$ with no
gradient information; they model unstructured input degradation.

\begin{figure*}[t]
  \centering
  \includegraphics[width=\linewidth]{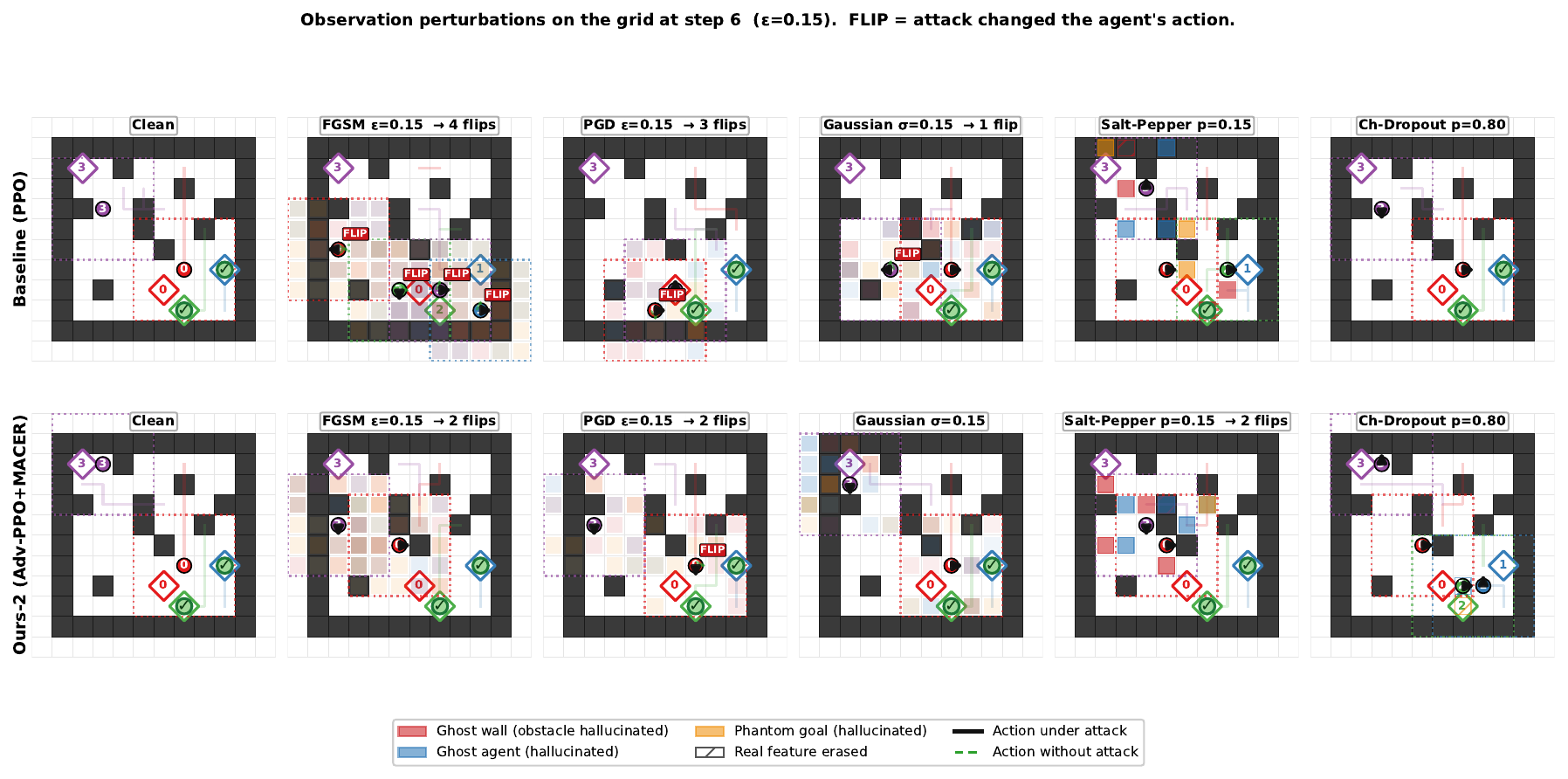}
  \caption{%
    \textbf{Six attack types visualized on the 8$\times$8 grid at step~6
    ($\epsilon{=}0.15$).}
    Top row: baseline PPO policy.  Bottom row: our Adv-PPO+MACER policy.
    Each agent's 5$\times$5 field of view is shown with a dashed border.
    Inside the field of view, \textcolor[HTML]{cb181d}{\textbf{red}} fill
    indicates a ghost wall the policy incorrectly sees (obstacle channel
    hallucinated); \textcolor[HTML]{2171b5}{\textbf{blue}} fill
    indicates a ghost agent; \textcolor[HTML]{f08c00}{\textbf{amber}} fill
    indicates a phantom goal hint.
    Hatched cells mark real features that have been erased.
    The solid black arrow shows the action actually taken under the attack;
    the dashed green arrow shows what the policy would have done on clean
    observations.  \textbf{FLIP} badges count action changes caused by the
    attack.  The baseline accumulates more flips under FGSM and PGD,
    demonstrating its vulnerability to adversarial perturbations.
    Our method consistently reduces the number of flipped actions.
  }%
  \label{fig:noise_comparison}
\end{figure*}

\paragraph{Objective and metrics.} Our goal is a shared policy $\pi_\theta$ that, for
every $o^{(i)}_t$ along the trajectory and for every attack
$\delta^{(i)}_t \in \mathcal{D}(o^{(i)}_t,\epsilon)$, still guides the team to its goals.
We measure this with two complementary metrics.

\noindent\emph{Empirical robustness}: the average success rate
$\mathrm{Succ}(\pi_\theta;a,\epsilon)$ under attack $a$ at budget
$\epsilon$, averaged over a held-out seed pool. Success of an episode
is the fraction of agents that reach their assigned goal within the
horizon. We report a per-cell value, the mean across attacked cells,
and the worst single attacked cell.

\noindent\emph{Certified action stability}: at a sampled state
$o^{(i)}_t$, randomized smoothing \citep{cohen2019} gives an
$\ell_2$-radius $R(o)$ such that the smoothed policy's argmax action is
provably unchanged for any perturbation of size at most $R(o)$. We
report $\bar{R}$, the mean of $R$ over a pool of trajectory states. The
certificate is for the smoothed policy, not the deployed unsmoothed
policy. We use the certificate as a sanity check on the smoothness of
the learned policy and we say so wherever we use it.

\paragraph{Scope.} This paper defends only the observation tensor. We do
not add any preprocessing between the sensor and the policy, and we do not
change the shared backbone architecture. The whole defense lives in the
training objective.

\paragraph{Notation summary.} Several Greek letters appear in the method
section with different meanings, so we list them here to avoid confusion.
$\epsilon$ is the evaluation attack budget; $\epsilon_{\text{train}}$ is
the budget used during training (these can differ); $\epsilon_s$ is the
radius of the uniform smoothness noise; $\epsilon_c$ is the PPO clip
width. $\sigma$ is the Gaussian noise standard deviation for randomized
smoothing. $\alpha_{\text{step}}$ is the PGD step size; $\alpha_{\text{adv}}$
is the fraction of each minibatch that gets adversarial inputs (a mixing
ratio). $\gamma_{\text{disc}}$ is the discount factor; $\gamma_M$ is the
target margin in the MACER hinge. $\lambda_{\text{GAE}}$ is the
generalized advantage estimation parameter; $\lambda_M$ is the weight
on the MACER term. $f_w$ and $f_r$ are the warm-up and ramp fractions
of the SA-PPO regularizer schedule; $\kappa^\star$ is its ceiling value.
$\phi$ denotes the source network used to generate an attack.

% =====================================================================
\section{Method}
\label{sec:method}
% =====================================================================

We build the recipe in two layers. Sec.~\ref{sec:advppo} introduces
Adv-PPO, which trains a shared policy to act consistently across
nearby observations. Sec.~\ref{sec:macer} stacks an on-policy
certified-radius hinge on top of that to obtain Adv-PPO+MACER.

\subsection{Adv-PPO: state-adversarial PPO with smoothness and a robust selector}
\label{sec:advppo}

Plain PPO maximizes return on the
clean trajectory distribution. It places no constraint on what the
policy does when the same observation is shifted by a small bounded
amount. The Adv-PPO loss adds three components that together make the
trainer care about a small neighbourhood of every observation:
training-time attacked inputs, a state-adversarial regularizer, and a
local smoothness regularizer. We then describe a checkpoint selector
that picks the released model by adversarial performance rather than
clean reward.

\paragraph{Training-time attacked inputs.} On a fraction
$\alpha_{\text{adv}}$ of each minibatch we replace the clean
observation $o_t$ with a perturbed one $\tilde{o}_t = o_t + \delta_t$.
The perturbation $\delta_t$ is computed by FGSM or PGD against a
\emph{frozen} pre-trained baseline $\theta^{\text{base}}$. We freeze
the source because, in early experiments where the source moved with
the defender, the policy entropy collapsed below $0.2$ nats within a
few iterations: when the attacker chases the defender, the defender
keeps moving away, and the only stable point is a deterministic
policy. A frozen source gives the defender a fixed adversary to
generalize against. The PPO importance ratio is recomputed at
$\tilde{o}_t$ in both numerator and denominator so the ratio remains
finite without an explicit correction.

\paragraph{State-adversarial regularizer (SA-PPO).} The
state-adversarial PPO regularizer asks the policy to produce nearly
the same action distribution at $o$ and at the worst $o+\delta$
inside the budget:
\begin{equation}
\begin{aligned}
R_{\text{SA}}(\theta) = \mathbb{E}_{o\sim d_{\pi_\theta}}\Big[
&\max_{\delta\in\mathcal{D}(o,\epsilon_{\text{train}})} \\[-1pt]
&D_{\mathrm{KL}}\big(\pi_{\bar\theta}(\cdot\mid o)\,\|\,
\pi_\theta(\cdot\mid o+\delta)\big) \Big].
\end{aligned}
\label{eq:sa-kl}
\end{equation}
Here $\bar\theta=\mathrm{stopgrad}(\theta)$ supplies a fixed
``teacher'' distribution at the clean input, and the inner $\max$
asks for the perturbation in the budget that most changes the
predicted distribution. Solving the inner $\max$ exactly cannot be
done in closed form, so we approximate it by sign-gradient PGD on $\delta$ for
$K_{\text{in}}$ steps. The intuition is simple: every gradient step
on this term shrinks the worst-case KL gap inside the
$\epsilon_{\text{train}}$-ball. \citet{sa-mdp} prove that, for a Lipschitz-bounded
policy, this gap upper-bounds the worst-case value gap between the
clean and the attacked policy. We use this regularizer because it is
the gradient of an explicit value-based bound, not the gradient of a
generic heuristic.

\paragraph{Local smoothness regularizer.} The SA-PPO term cares about
the worst $\delta$ in the ball. A second, cheaper term cares about
the average $\delta$ in the ball. Sampling a uniform perturbation
$\delta\sim\mathcal{U}(-\epsilon_s,\epsilon_s)$ and matching the
distribution at $o$ to the distribution at $o+\delta$ acts as a
local-Lipschitz prior on $\pi_\theta$:
\begin{equation}
\begin{aligned}
R_{\text{TR}}(\theta) = \mathbb{E}\Big[
D_{\mathrm{KL}}\big(&\pi_{\bar\theta}(\cdot\mid o)\,\| \\[-1pt]
&\pi_\theta(\cdot\mid o+\delta)\big) \Big],
\end{aligned}
\label{eq:trades}
\end{equation}
with the expectation taken over $o\sim d_{\pi_\theta}$ and
$\delta\sim\mathcal{U}(-\epsilon_s,\epsilon_s)$. This is a
single-sample relative of TRADES \citep{trades} that drops the inner
PGD step. It is cheap to compute (one extra forward pass per
minibatch) and complements the SA-PPO term by penalizing curvature
even at non-adversarial perturbations.

\paragraph{Putting the pieces together.} Combining the three terms
with the standard PPO surrogate, value loss, and entropy bonus gives
the Adv-PPO objective
\begin{align}
\mathcal{L}(\theta) =\;& \mathcal{L}_{\text{PPO}}(\theta;\tilde o)
+ c_v\,\mathcal{L}_V(\theta;\tilde o) \notag\\
&- c_H\,H[\pi_\theta(\cdot\mid\tilde o)]
+ \beta\,R_{\text{TR}}(\theta) \notag\\
&+ \kappa(n)\,R_{\text{SA}}(\theta).
\label{eq:full-loss-advppo}
\end{align}
The schedule $\kappa(n)$ on the SA-PPO term holds it at zero for the
first fraction $f_w$ of training, then ramps it linearly to
$\kappa^\star$ over the next fraction $f_r$, then keeps it constant.
The reason is staging: the policy first needs to learn the task on
clean inputs, and only then can it pay the cost of a smoothness
constraint without losing the task signal.

\paragraph{Robust checkpoint selection.} The released model is not
the iteration with the highest clean reward but the iteration with
the highest score on a held-out attacked validation pool. Every
$T_{\text{eval}}$ outer iterations we evaluate the current $\theta$
under FGSM and PGD at $\epsilon\in\{0.10,0.20\}$ and define
\begin{equation}
\mathrm{Score}(\pi)=\tfrac14\!\!\sum_{a\in\{\text{fgsm,pgd}\}}\!\!
\sum_{\epsilon\in\{0.10,0.20\}}\mathrm{Succ}(\pi;a,\epsilon).
\label{eq:score}
\end{equation}
We pick the iteration with the highest $\mathrm{Score}$ over the
training run. In our runs, picking by clean training success
typically returned an early iteration that was somewhat fragile;
attacked-set selection consistently returned a later iteration with
slightly worse clean reward and substantially better worst-case
behaviour.

\subsection{Adv-PPO+MACER: an on-policy certified-smoothness hinge}
\label{sec:macer}

\paragraph{Certified-radius hinge (MACER).} The SA-PPO term constrains
the policy on the worst $\delta$ inside the $\ell_\infty$ ball at
each visited state. It does not say how confident the policy should
be in its top action under noise, only that the top action should not
flip. Randomized smoothing offers a complementary signal. For an
isotropic Gaussian smoothing of width $\sigma$, the smoothed
top-action probability $p_A(o)$ gives a strict $\ell_2$ stability
radius
\[
R(o)=\sigma\,\Phi^{-1}(p_A(o)),
\]
where $\Phi^{-1}$ is the inverse standard-normal cumulative distribution function (CDF)
\citep{cohen2019}. Inside that radius the smoothed policy's argmax
does not change. MACER \citep{macer} writes a differentiable hinge
that pushes $R(o)$ above a target margin $\gamma_M$. Adding this
hinge to the PPO loss therefore gives the trainer a direct gradient
on the certified-radius lower bound, which is the most explicit
margin signal available for a smoothed classifier. We chose MACER
over SmoothAdv (which requires an inner adversarial PGD on
Gaussian-perturbed inputs and roughly doubles the training cost) and
over interval-bound propagation (which is loose for ReLU CNNs of the
size used here) because MACER fits a shared convolutional neural network (CNN) with no architectural
change and adds one extra forward pass per minibatch.

\paragraph{The MACER hinge.} For each sampled state $o$ in the batch
we draw $K_M$ Gaussian samples $\xi_k\sim\mathcal{N}(0,\sigma^2 I)$,
evaluate $\pi_\theta(\cdot\mid o+\xi_k)$, and form the empirical
smoothed distribution. Let $A^\star(o)=\arg\max_a\pi_\theta(a\mid o)$
be the clean top action. The empirical top frequency at $A^\star$ and
the empirical runner-up frequency over the smoothed samples are
\[
\hat p_A(o)=\tfrac{1}{K_M}\!\!\sum_{k=1}^{K_M}\!\!\mathbf{1}\{\arg\max_a\pi_\theta(a\mid o+\xi_k)=A^\star(o)\},
\]
and $\hat p_B(o)$, where $B$ is the most-frequent action under the
smoothed samples other than $A^\star(o)$. Define the empirical
smoothed argmax $\tilde{A}(o)=\arg\max_a \hat p_a(o)$. Following the
binary form of \citet{macer},
\begin{equation}
\begin{aligned}
\mathcal{L}_M(\theta;o) = \mathbf{1}\{\tilde{A}(o)=A^\star(o)\}\cdot \max\Big(0,\;
\gamma_M - \\[-1pt]
\tfrac{\sigma}{2}\big(\Phi^{-1}(\hat p_A) -\Phi^{-1}(\hat p_B)\big)\Big),
\end{aligned}
\label{eq:macer-hinge}
\end{equation}
where the indicator masks the hinge to the subset of states where the
empirical smoothed argmax already matches the clean argmax ($\tilde{A}(o)$ uses tilde to distinguish from the GAE advantage estimate $\hat{A}_t$ used in the algorithm). The mask
matters: without it, the hinge would penalize states where the
smoothed policy is already wrong, which destabilizes the RL signal.
The hinge is zero whenever the smoothed margin already exceeds
$\gamma_M$, and is linear in the deficit otherwise.

\paragraph{Two-step proximal optimization.} The PPO surrogate clip
keeps the policy ratio bounded above by a small constant; the MACER
hinge has no such bound and can be much larger when the smoothed
margin is small. Adding the two losses with a single combined
backward pass made training unstable in our runs because the two loss
surfaces have very different scales. We therefore apply the MACER
term as a small proximal step on the same minibatch immediately after
the standard Adv-PPO step:
\begin{equation}
\begin{aligned}
\theta &\leftarrow \theta - \eta\,\nabla_\theta\,\mathcal{L}_{\text{Adv-PPO}}(\theta), \\
\theta &\leftarrow \theta - \eta\,\lambda_M\,\nabla_\theta\,
\mathbb{E}_{o\sim d_{\pi_\theta}}\!\bigl[\mathcal{L}_M(\theta;o)\bigr].
\end{aligned}
\label{eq:full-loss-macer}
\end{equation}
Two design choices need to hold for this to work in PPO. First,
$\lambda_M$ must be small. With a large $\lambda_M$ the cheapest way
to satisfy the hinge is to push the top-action probability to one at
every visited state, which removes the entropy PPO needs in order to
keep exploring; we use $\lambda_M=0.05$. Second, the entropy
coefficient must be slightly increased to counter-balance the hinge,
which would otherwise push the policy toward a deterministic mode;
we use $c_H=0.05$ instead of the standard PPO default $c_H=0.01$.

\paragraph{On-policy versus post-hoc.} The natural alternative is to
first train an Adv-PPO teacher and then distill it into a smoothed
student by minimizing $\mathcal{L}_M$ on the teacher's
trajectories. This produces the largest certified radius in our
experiments ($\bar R=0.207$) but, as we report in
Sec.~\ref{sec:results}, drops the deployed clean success rate from
$96.7\%$ to $57.5\%$. The reason is the same as the one DAGGER
\citep{dagger} addresses in imitation learning: the post-hoc student
acts under smoothing and visits states the teacher never labelled, so
the supervised signal is silent on the shifted state distribution. The
on-policy variant in Eq.~\eqref{eq:full-loss-macer} avoids this
because the hinge is computed on the policy's own current trajectory
distribution, so the smoothness signal is always paired with the
states the policy will actually see at deployment.

\begin{algorithm}[t]
\caption{One outer iteration of Adv-PPO+MACER. For Adv-PPO alone, omit the MACER step (line~17); everything else is identical.}
\label{alg:advppo-macer}
\begin{algorithmic}[1]
\REQUIRE Defender $\theta$, frozen baseline $\theta^{\text{base}}$,
schedule $\kappa(n)$ with parameters $f_w,f_r,\kappa^\star$, PPO
hyperparameters $c_v,c_H,\epsilon_c$, robust-selector period
$T_{\text{eval}}$, attack hyperparameters
$\alpha_{\text{adv}},\beta,\epsilon_{\text{train}},\epsilon_s,K_{\text{in}}$, MACER
hyperparameters $\lambda_M,\sigma,\gamma_M,K_M$.
\STATE Roll out a batch of episodes with $\pi_\theta$ on clean
observations; collect $\{o_t,a_t,\hat A_t,R_t\}$, with $\hat A_t$ the
GAE advantage \citep{gae} (Generalized Advantage Estimation) and $R_t$ the discounted return at step $t$.
\FOR{each PPO epoch}
  \FOR{each minibatch $\mathcal{B}$}
    \STATE Sample $M_{\text{adv}}\subset\mathcal{B}$ with i.i.d.
    probability $\alpha_{\text{adv}}$.
    \STATE For $t\in M_{\text{adv}}$, build $\delta_t$ by FGSM or PGD
    on $\theta^{\text{base}}$ targeting $a^{\text{clean}}_t$ and form
    $\tilde o_t=o_t+\delta_t$; for $t\notin M_{\text{adv}}$, set
    $\tilde o_t=o_t$.
    \STATE Recompute $\pi_{\theta_{\text{old}}}(a_t\mid \tilde o_t)$.
    \STATE Compute $\mathcal{L}_{\text{PPO}}$, $\mathcal{L}_V$,
    $H[\pi_\theta]$ at $\tilde o_t$.
    \STATE Compute $R_{\text{TR}}$ from Eq.~\eqref{eq:trades}.
    \IF{$\kappa(n)>0$}
      \STATE Compute $R_{\text{SA}}$ from Eq.~\eqref{eq:sa-kl} via
      $K_{\text{in}}$-step PGD on $\delta$.
    \ELSE
      \STATE $R_{\text{SA}}\leftarrow 0$.
    \ENDIF
    \STATE \emph{Adv-PPO step:} update $\theta$ on
    Eq.~\eqref{eq:full-loss-advppo}; clip gradient.
    \STATE \emph{MACER step (Adv-PPO+MACER only):} estimate
    $\hat p_A(o),\hat p_B(o)$ from $K_M$ Gaussian samples on the same
    $o$; compute $\mathcal{L}_M$ from Eq.~\eqref{eq:macer-hinge};
    update $\theta$ on $\lambda_M\,\mathcal{L}_M$.
  \ENDFOR
\ENDFOR
\STATE Every $T_{\text{eval}}$ iterations, evaluate
$\mathrm{Score}(\pi_\theta)$ from Eq.~\eqref{eq:score}; keep best.
\end{algorithmic}
\end{algorithm}

% =====================================================================
\section{Experimental Setup}
% =====================================================================

\paragraph{Environment.} POGEMA \citep{pogema} on grids of side $L=8$
with obstacle density $\rho=0.1$, $N=4$ agents per episode, observation
radius $r=2$ (so each $o^{(i)}_t\in\{0,1\}^{3\times5\times5}$), and a
horizon of $T=64$ steps. Random map, agent starts, and goals are drawn
fresh per seed. The training, validation, and reporting seed pools are
disjoint.

\paragraph{Backbone.} A two-layer CNN with $32$ then $64$ channels, a
$128$-unit linear trunk, a $5$-way actor head, and a scalar critic
head. The same backbone is used for the baseline, both proposed
methods, and the post-hoc baselines, so any robustness gap is
attributable to the training objective and not to capacity.

\paragraph{Baseline (PPO).} Trained with shared-parameter PPO on the
same environment-step budget as the defenders. Clean training success
plateaus near $96\%$. The baseline checkpoint is then frozen and used
both as the source of training-time attacks and as the initialization
for the two proposed methods.

\paragraph{Adv-PPO.} Adam at $3\!\times\!10^{-4}$, GAE
$\lambda_{\text{GAE}}=0.95$, discount $\gamma_{\text{disc}}=0.95$, value coefficient
$c_v=0.5$, entropy coefficient $c_H=0.01$, PPO clip width
$\epsilon_c=0.2$, four PPO epochs per outer iteration, four
minibatches. $\alpha_{\text{adv}}=0.30$, training-time attack budget
$\epsilon_{\text{train}}=0.15$, smoothness penalty $\beta=0.80$ at
$\epsilon_s=0.08$. The SA-PPO inner loop uses $K_{\text{in}}=5$ steps
at $\epsilon_{\text{train}}=0.15$; $\kappa^\star=0.80$. The SA schedule uses warm-up
fraction $f_w=0.05$ and ramp fraction $f_r=0.15$. Robust
checkpoint selection uses $T_{\text{eval}}=4$, eight evaluation
episodes per cell.

\paragraph{Adv-PPO+MACER.} We initialize from the strongest available
Adv-PPO checkpoint and fine-tune at Adam learning rate
$5\!\times\!10^{-5}$ for $50{,}000$ environment steps with
$\lambda_M=0.05$, $\sigma=0.10$, $\gamma_M=0.20$, $K_M=4$, $c_H=0.05$,
$\alpha_{\text{adv}}=0.40$, MACER warm-up fraction $0.20$. We repeat
the fine-tune for three independent seeds (42, 7, 123). The upstream
PPO and Adv-PPO checkpoints come from a single training seed; we
discuss the implication in Sec.~\ref{sec:limitations}.
Note: the training script CLI defaults for $\lambda_M$ and $\gamma_M$
differ from these values; the paper values were passed explicitly as
command-line flags during each reported run.

\paragraph{Evaluation protocol.} For each of $21$ attack settings
(FGSM at five $\epsilon$ values, PGD at five $\epsilon$ values,
Gaussian sensor noise at four $\sigma_{\text{sens}}$ values,
salt-and-pepper at four rates, channel dropout at three rates;
$5{+}5{+}4{+}4{+}3=21$) we run $30$ episodes per setting on a fixed
seed pool that was not touched during training or selection. Each
episode in cell $(a,\epsilon,k)$ uses the deterministic seed
$50000+13\cdot k+7\cdot n_{\text{adv}}$, so different methods are
compared on identical worlds. PGD uses $K=10$ steps with
$\alpha_{\text{step}}=2\epsilon/K$ from one random initialization in
$\mathcal{D}$; we use no additional restarts. We report mean
success across episodes per cell, the average of the per-cell values
across all attacked settings, and the single worst attacked cell. The
certified radius is computed via Cohen smoothing
\citep{cohen2019} with $\sigma=0.10$, $n_0=32$ selection samples,
$n=500$ estimation samples, $\alpha_{\text{conf}}=10^{-3}$ confidence
level, on a pool of $1500$ trajectory states. Cross-seed standard
deviations for Ours-2 use sample standard deviation (ddof $=1$).
The certification evaluation defaults in code use smaller pool sizes;
the values above ($n{=}500$, $N{=}1500$) were passed as explicit flags.

% =====================================================================
\section{Results}
\label{sec:results}
% =====================================================================

\subsection{Headline numbers}

Table~\ref{tab:headline} reports the headline numbers across five
methods on the same seeds and the same evaluation script. The
unprotected PPO baseline reaches $95.8\%$ clean success but only
$2.5\%$ in the worst attacked cell. Adv-PPO (Ours-1) reaches $59.2\%$
worst-case at no clean cost ($96.7\%$ vs $95.8\%$). Adv-PPO+MACER
(Ours-2) reaches $77.5\%\pm 6.0\%$ worst-case across three independent
seeds, again at near-zero clean cost ($95.0\pm 0.8\%$). A paired
bootstrap over the $21$ attack cells (10{,}000 resamples) gives a
mean adversarial-success gap of Ours-2 over Ours-1 of $+2.4$
percentage points with $95\%$ CI $[+0.4,+4.9]$, so the gain from the
MACER hinge is positive across the cell distribution and not driven
by a single cell.

\begin{table}[t]
\centering
\footnotesize
\setlength{\tabcolsep}{3pt}
\resizebox{\columnwidth}{!}{%
\begin{tabular}{lcccc}
\toprule
Method & Clean (\%) & Mean adv. (\%) & Worst adv. (\%) & Cert.\ $\bar{R}$ \\
\midrule
PPO & 95.8 & 73.3 & 2.5 & 0.088 \\
MACER (post-hoc) & 59.2 & 65.0 & 8.3 & 0.183 \\
Adv-PPO+MACER (post-hoc) & 57.5 & 61.4 & 0.8 & 0.207 \\
Adv-PPO (Ours-1) & 96.7 & 92.4 & 59.2 & 0.139 \\
Adv-PPO+MACER (Ours-2) & 95.0\,$\pm$\,0.8 & 94.8\,$\pm$\,0.3 & 77.5\,$\pm$\,6.0 & 0.141\,$\pm$\,0.016 \\
\bottomrule
\end{tabular}%
}
\caption{Robustness on POGEMA MAPF (8\,$\times$\,8 grid, 4 agents, 30 episodes per attack setting, 21 attack cells). Bands for Ours-2 are sample standard deviation (ddof $=1$) across three independent fine-tune seeds (42, 7, 123); the upstream PPO and Adv-PPO checkpoints are single-seed.}
\label{tab:headline}
\end{table}

The two post-hoc rows are the negative result. They obtain the largest
certified radius in our table ($\bar R=0.207$ for the post-hoc
combined method, vs $0.141$ for Ours-2) but their deployed clean
success drops below $60\%$. They are unsafe to deploy as policies even
though they look strong as classifiers.

\begin{figure}[t]
\centering
\includegraphics[width=\columnwidth]{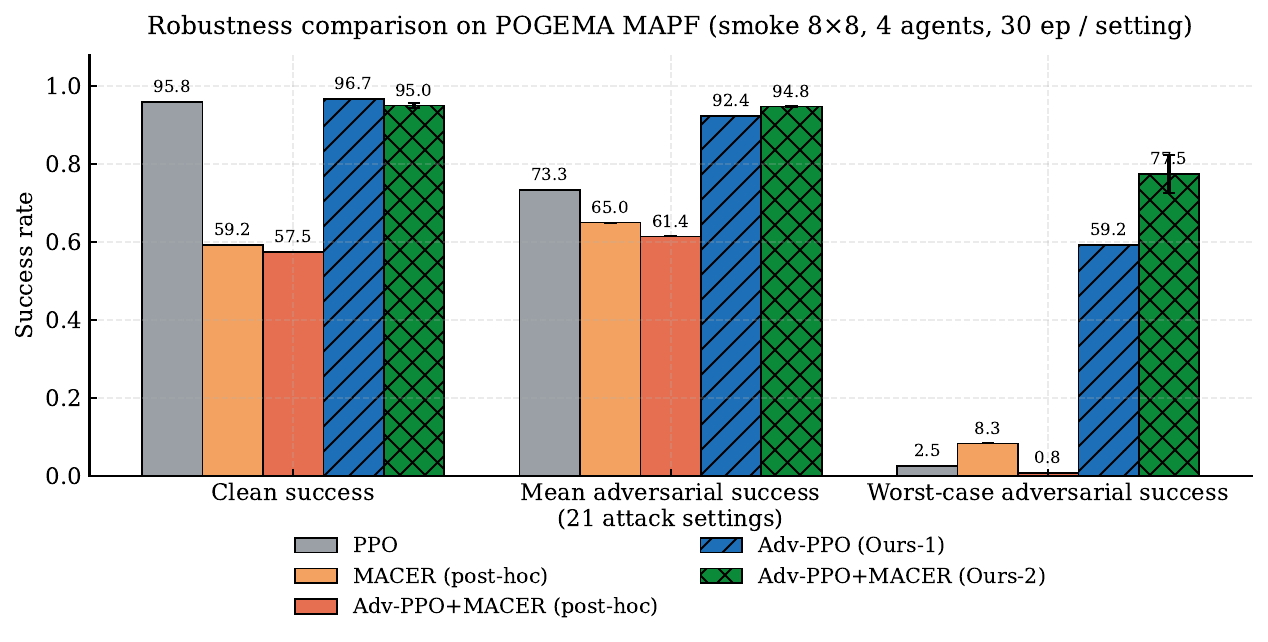}
\caption{Headline comparison on POGEMA $8{\times}8$, $4$ agents, $30$
episodes per attack setting. Each method reports clean success, the
mean attacked success across $21$ attack settings, and the worst single
attacked cell. The two proposed methods (Ours-1 and Ours-2) keep clean
performance and recover most of the worst-case loss; the two post-hoc
rows show the negative result.}
\label{fig:headline}
\end{figure}

\begin{figure}[t]
\centering
\includegraphics[width=\columnwidth]{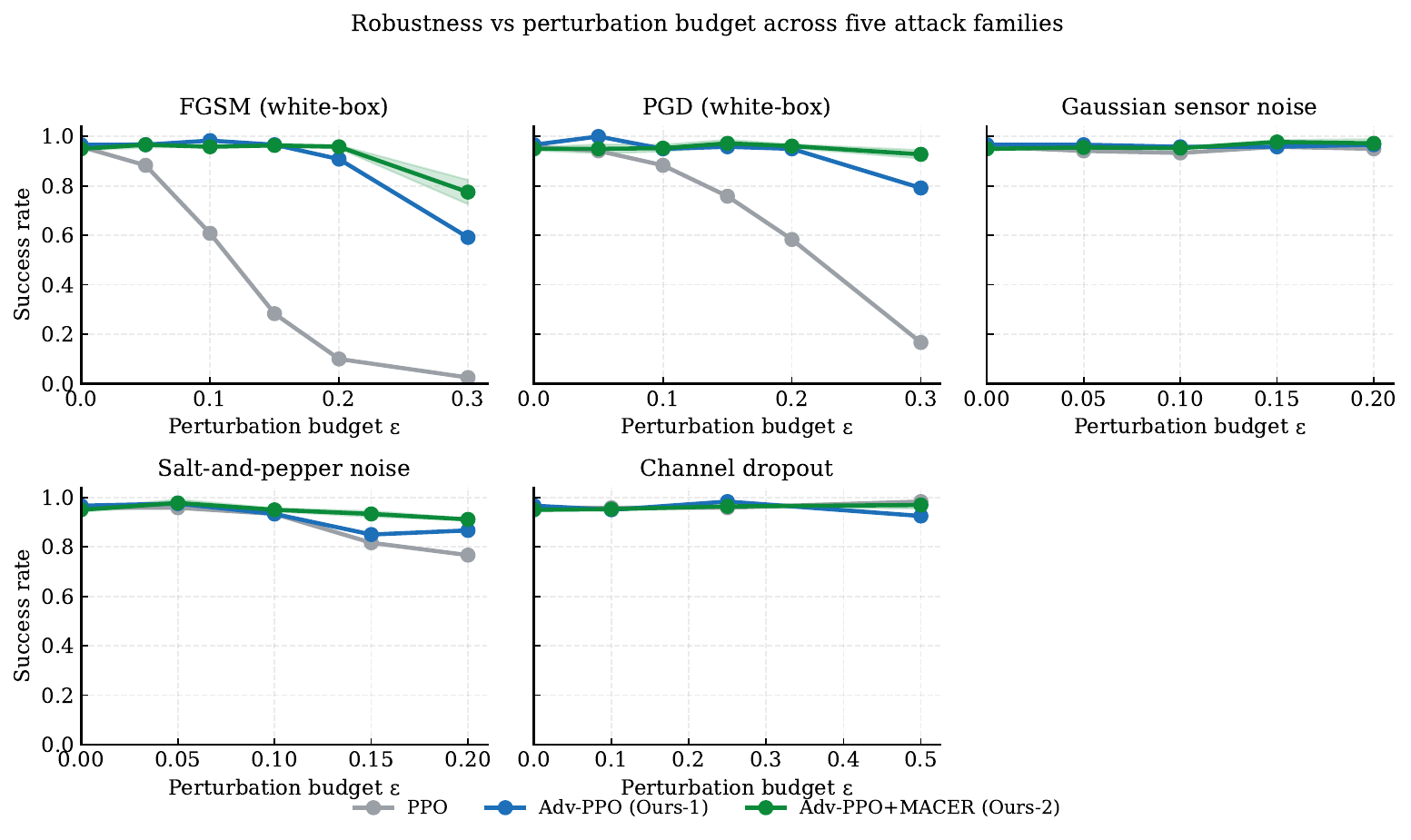}
\caption{Success rate vs.\ attack budget for three of the five
attack families (FGSM, PGD, Gaussian sensor noise). Three methods are
shown for readability (PPO baseline, Ours-1, Ours-2); the two post-hoc
rows of Table~\ref{tab:headline} are omitted from this figure because
their clean rate is below $60\%$ on every attack budget and they
overlap visually. Adv-PPO+MACER (Ours-2) tracks Adv-PPO (Ours-1) in
every cell and is strictly above the unprotected PPO baseline.}
\label{fig:per-attack}
\end{figure}

Figure~\ref{fig:headline} plots the same numbers as a grouped bar
chart over Clean, Mean adversarial, and Worst adversarial success.
Figure~\ref{fig:per-attack} expands the mean-over-21 number into the
full per-attack curves; Adv-PPO+MACER stays close to its clean
success in every cell, including under the strongest PGD setting.

\subsection{Certified action stability}

\begin{figure}[t]
\centering
\includegraphics[width=\columnwidth]{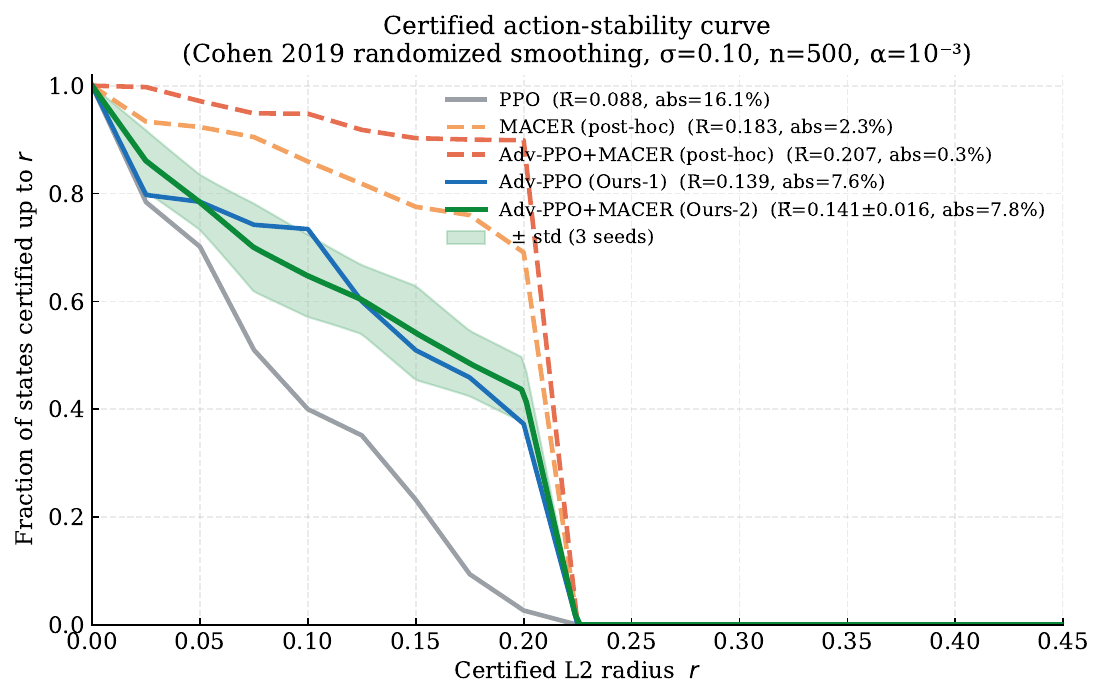}
\caption{Certified action-stability curve from randomized smoothing
\citep{cohen2019} at $\sigma=0.10$, $n=500$, $\alpha_{\text{conf}}=10^{-3}$, on
$1500$ trajectory states. The post-hoc smoothing baselines maximize
this certificate but at a deployed-clean cost we describe in the text.
The shaded band on Ours-2 is the sample standard deviation (ddof $=1$)
across three seeds; the exact value of $\bar R$ for Ours-2 is reported
in Table~\ref{tab:headline}.}
\label{fig:cert}
\end{figure}

Figure~\ref{fig:cert} reports the certified action-stability curve for
the same five methods. The post-hoc smoothing baselines have the
largest certified radius ($\bar R=0.207$) but with the deployed-policy
collapse described above. The two proposed methods give a slightly
smaller certified radius than that, and a substantially larger one
than the PPO baseline. The certificate is a property of
the smoothed policy, not the deployed unsmoothed policy. Wrapping any
of these policies in the Cohen smoothing classifier at $\sigma=0.10$
drops clean success to about $65\%$ across the board, so we report the
certificate as a smoothness sanity check on the trained policy and
deploy the unsmoothed model.

\subsection{Clean-vs-robust trade-off}

\begin{figure}[t]
\centering
\includegraphics[width=\columnwidth]{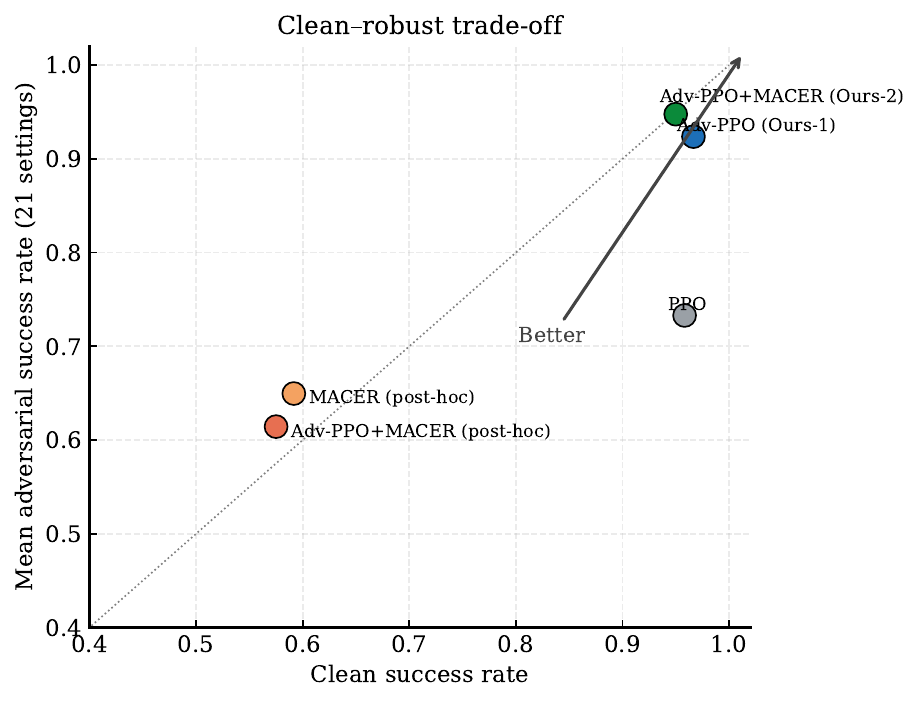}
\caption{Clean success rate (x-axis) vs.\ mean attacked success rate
across $21$ attack settings (y-axis) for the five methods. Bigger is
better on both axes. The two proposed methods sit in the top-right
corner.}
\label{fig:tradeoff}
\end{figure}

Figure~\ref{fig:tradeoff} places each method on the
clean-vs-mean-attacked plane. The post-hoc methods sit in the
top-left corner: high robustness, low clean success, low joint score.
The two proposed methods sit in the top-right corner: a method is good
on this axis only if it is good on both.

\subsection{Rollout storyboards}

\begin{figure*}[tb]
\centering
\includegraphics[width=0.95\textwidth]{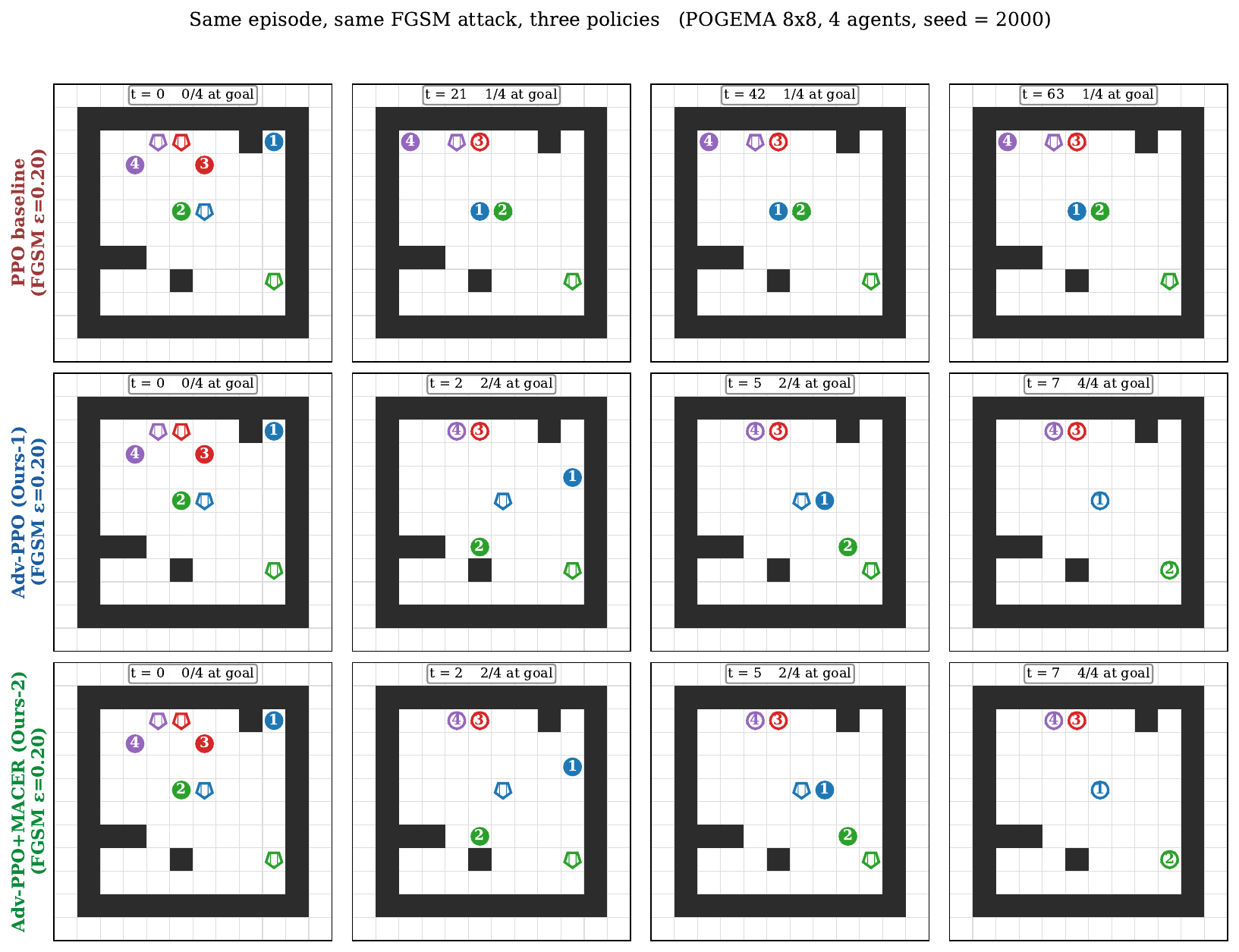}
\caption{Rollout storyboard on a single POGEMA episode (seed $2000$)
under FGSM at $\epsilon=0.20$, for three policies.
\emph{Top:} PPO baseline --- agents move erratically and reach only
one of four goals within the horizon.
\emph{Middle:} Adv-PPO (Ours-1) --- all four goals reached.
\emph{Bottom:} Adv-PPO+MACER (Ours-2) --- all four goals reached.
Start cells, goal cells, obstacles, and the random seed are identical
across all three rows; only the training objective differs.
Goal cells are drawn as outlined markers; agents are filled circles
labelled by index; an open white circle marks an agent that has
reached its goal.}
\label{fig:rollout-three}
\end{figure*}

Numbers alone can be hard to interpret.
Figure~\ref{fig:rollout-three} translates the headline results into
actual agent trajectories, showing the same POGEMA episode (seed
$2000$, identical start and goal cells) under FGSM at $\epsilon=0.20$
for all three policies side by side.
The PPO baseline (top row) moves erratically under the perturbed
observations and reaches only one of four goals within the horizon.
Adv-PPO (middle row) completes all four goals, showing that
adversarial training alone is enough to recover most robustness.
Adv-PPO+MACER (bottom row) also completes all four goals; the
improvement over Adv-PPO shows up in the cross-seed numbers
(Table~\ref{tab:headline}) rather than on this single episode.

\subsection{Cross-seed reproducibility for Ours-2}

The cross-seed numbers for Adv-PPO+MACER are: mean attacked success
$94.6\%/95.1\%/94.6\%$ on seeds $42/7/123$ and worst attacked cell
$72.5\%/75.8\%/84.2\%$. All three seeds beat Adv-PPO on both metrics.
Sample standard deviations (ddof $=1$) across the three seeds are
reported in Table~\ref{tab:headline} and as the shaded band on Ours-2
in Fig.~\ref{fig:cert}. We report these explicitly because robustness
numbers in deep RL are well known to move with the seed; with three
seeds the worst-case gap over the strongest non-MACER baseline is
robust to the seed choice.

\subsection{Multi-restart PGD sanity check}
\label{sec:mr-pgd}

A single random PGD initialization can underestimate worst-case
behaviour. To check that the ranking in
Table~\ref{tab:headline} is not an artifact of weak attack search, we
re-evaluate the three deployed policies (PPO baseline, Adv-PPO,
Adv-PPO+MACER) under PGD with five independent random
initializations per episode and take the minimum success rate across
the five restarts. The protocol is otherwise identical to
Sec.~\ref{sec:method}: $K=10$ PGD steps per restart,
$\alpha_{\text{step}}=2\epsilon/K$, $30$ episodes per
restart on the same seed pool.
Table~\ref{tab:mr-pgd} reports the worst-of-five
success rate at $\epsilon\in\{0.20,0.30\}$. The ranking is preserved
at both budgets: Adv-PPO+MACER $>$ Adv-PPO $>$ PPO. The drops from
$1$-restart to $5$-restart are modest ($3$ to $8$ percentage points),
and the absolute gap of Ours-2 over the unprotected baseline is
$10$ and $33$ percentage points at $\epsilon=0.20$ and $\epsilon=0.30$
respectively, an order of magnitude larger than the multi-restart
correction. We therefore continue to report $1$-restart numbers in
Table~\ref{tab:headline} for comparability with prior work and treat
Table~\ref{tab:mr-pgd} as a robustness check on the conclusion.

\begin{table}[t]
\centering
\resizebox{\columnwidth}{!}{%
\begin{tabular}{lcc}
\toprule
Method & PGD-5 @\,$\epsilon{=}0.20$ & PGD-5 @\,$\epsilon{=}0.30$ \\
\midrule
PPO baseline      & 72.5 & 42.5 \\
Adv-PPO (Ours-1)  & 79.2 & 68.3 \\
Adv-PPO+MACER (Ours-2) & \textbf{82.5} & \textbf{75.8} \\
\bottomrule
\end{tabular}%
}
\caption{Worst-of-five-restart PGD success rate (\%) on POGEMA
$8{\times}8$ with $4$ agents, $30$ episodes per restart,
$K{=}10$ PGD steps per restart. Five independent random
initializations per episode; reported value is the minimum success
across restarts. The ranking from Table~\ref{tab:headline} is
preserved at both budgets.}
\label{tab:mr-pgd}
\end{table}

% =====================================================================
\section{Discussion}
\label{sec:discussion}
% =====================================================================

The post-hoc distillation rows in Table~\ref{tab:headline} achieve
near-perfect training accuracy yet fail at deployment because the
smoothed student visits states that the Adv-PPO teacher never
encountered, leaving the smoothness signal uninformative on those
states. Our on-policy Adv-PPO+MACER variant avoids this: the
smoothness loss is computed on the policy's own current trajectories,
so the signal is always relevant to what the agent will see at test
time. For the MACER weight, we found that $\lambda_M \ge 0.30$
collapses policy entropy within a few iterations; $\lambda_M=0.05$
keeps exploration healthy while still tightening the worst-case
success. Finally, the certified radius $\bar{R}$ in
Table~\ref{tab:headline} belongs to a Cohen-smoothed wrapper, not
the deployed argmax policy — wrapping drops clean success to around
$65\%$, making it impractical for MAPF. We report it as a
training-time smoothness indicator and a method-to-method comparison
tool, not as a deployment guarantee. The practical motivation for the
$\ell_\infty$ threat model is that sensor errors, detection noise,
and reflection artefacts all fall inside a small per-pixel budget;
a policy that survives the worst-case gradient attack tends to
survive these unstructured variants as well, which the per-attack
results in Fig.~\ref{fig:per-attack} confirm.

% =====================================================================
\section{Limitations}
\label{sec:limitations}
% =====================================================================

\begin{itemize}
\item \textbf{Scale.} All results use an $8{\times}8$ grid with
$4$ agents. We have not verified the approach on larger maps or
larger teams; whether the gains hold at scale is an open question.
\item \textbf{Attack strength.} PGD uses $K{=}10$ steps from a single
random start. Stronger attack protocols may lower the reported numbers,
and we do not claim the current figures represent a true worst case.
\end{itemize}

% =====================================================================
\section{Conclusion}
% =====================================================================

A standard PPO MAPF policy on POGEMA loses almost all its success
under a small $\ell_\infty$ observation attack. We proposed two
training recipes that fix this without changing the network
architecture or the deployment pipeline. Adv-PPO ports SA-PPO to the
shared-policy multi-agent case and adds a smoothness regularizer and a
robust checkpoint selector; it lifts the worst-case attacked success
from $2.5\%$ to $59.2\%$ at no clean cost. Adv-PPO+MACER adds the
certified-radius hinge of MACER as an on-policy term with a small
weight and a matching entropy bonus; it lifts the worst-case to
$77.5\%\pm 6.0\%$ across three seeds. We also document an explicit
negative result (post-hoc MACER distillation collapses the deployed
policy) and explain why the on-policy variant avoids it. All numbers
are reproducible from the released code and the released seed pool.
The recipe is generic to shared-policy decentralized MARL: the SA-PPO
penalty, the TRADES smoothness term, and the MACER hinge make no
assumption about the MAPF structure beyond a per-agent local
observation, so the same pipeline could be applied to other
shared-policy benchmarks such as SMAC (StarCraft Multi-Agent Challenge) or MPE (Multi-Particle Environments).

% ----- Bibliography ----------------------------------------------------
% Note: aaai25.sty sets bibliographystyle automatically.
\bibliography{riad_paper}

% =====================================================================
% APPENDIX — Source Code
% =====================================================================
\clearpage
\onecolumn
\end{document}